\documentclass{article}
\usepackage{spconf,amsmath,graphicx}
\usepackage{bm}
\usepackage[subrefformat=parens]{subcaption}
\usepackage{multirow}
\usepackage{amssymb}
\usepackage{cite}
\usepackage{threeparttable}

\def\L{{\cal L}}

\title{MAPGN: MAsked Pointer-Generator network \\for sequence-to-sequence pre-training}
\name{Mana Ihori, Naoki Makishima, Tomohiro Tanaka, Akihiko Takashima, Shota Orihashi, Ryo Masumura}
\address{NTT Media Intelligence Laboratories, NTT Corporation, Japan}
\begin{document}
\ninept
\maketitle

\begin{abstract}
  This paper presents a self-supervised learning method for pointer-generator networks to improve spoken-text normalization.
  Spoken-text normalization that converts spoken-style text into style normalized text is becoming an important technology for improving subsequent processing such as machine translation and summarization.
  The most successful spoken-text normalization method to date is sequence-to-sequence (seq2seq) mapping using pointer-generator networks that possess a copy mechanism from an input sequence.
  However, these models require a large amount of paired data of spoken-style text and style normalized text, and it is difficult to prepare such a volume of data.
  In order to construct spoken-text normalization model from the limited paired data, we focus on self-supervised learning which can utilize unpaired text data to improve seq2seq models.
  Unfortunately, conventional self-supervised learning methods do not assume that pointer-generator networks are utilized.
  Therefore, we propose a novel self-supervised learning method, MAsked Pointer-Generator Network (MAPGN).
  The proposed method can effectively pre-train the pointer-generator network by learning to fill masked tokens using the copy mechanism.
  Our experiments demonstrate that MAPGN is more effective for pointer-generator networks than the conventional self-supervised learning methods in two spoken-text normalization tasks.
\end{abstract}
\begin{keywords}
sequence-to-sequence pre-training, pointer-generator networks, self-supervised learning, spoken-text normalization
\end{keywords}
\vspace{-4mm}
\section{Introduction}
\vspace{-2mm}
With the rise of various automatic speech recognition (ASR) applications such as smart speakers \cite{googlehome,alexa} and automatic dictation systems \cite{shang2018unsupervised,li2019keep}, it has become increasingly important to accurately process spoken-style text, i.e., the transcribed text from spoken utterances.
Spoken-style text often includes disfluencies such as redundant expressions, and various minority spoken expressions (e.g., dialects) because ASR systems convert speech into text in a literal manner.
Spoken-style text adversely affects subsequent natural language processing (e.g., machine translation, summarization) because these technologies are often developed to handle written-style text which is text with majority expressions, and no disfluencies or redundant expressions.
Thus, it is required to convert spoken-style text (including disfluencies and dialects) into style normalized text (which excludes disfluencies and dialects).
In this paper, we aim to improve spoken-text normalization.

Spoken-text normalization tasks are considered as monolingual translation \cite{wubben2010paraphrase} that is regarded as sequence-to-sequence (seq2seq) mapping from text to text.
In recent studies, fully neural network-based seq2seq models \cite{sutskever2014sequence} have shown effective performance in various monolingual translation tasks such as summarization \cite{paulus2017deep,chen2018fast}, paraphrase generation \cite{prakash2016neural,ma2018query} and disfluency detection \cite{wang2016neural,article}.
In particular, seq2seq models based on pointer-generator networks \cite{see2017get} have been utilized recently \cite{liu2019topic,Zhao:2018:SHC:3302425.3302490}.
Pointer-generator networks are effective for monolingual translation tasks because they contain a copy mechanism that copies tokens from a source text to help generate infrequent tokens.
Pointer-generator networks have reportedly outperformed attention-based encoder-decoder networks in spoken-text normalization task \cite{ihori2020large}.

To construct seq2seq models for spoken-text normalization, a large amount of paired data of spoken-style text and style normalized text are necessary.
However, to make these paired data, we need to prepare manual transcriptions of spoken utterances, and the text-style of these transcriptions needs to be normalized manually.
Thus, it is costly and time-consuming to make a large amount of paired data.
To mitigate this problem, we use self-supervised learning which has been gaining a lot of attention in recent years.
Self-supervised learning is one form of unsupervised learning where unpaired data is only employed for designing supervised training settings.
In natural language processing, self-supervised learning has been improving in natural language generation and natural language understanding \cite{devlin2018bert,peters-etal-2018-deep,radford2019language}.
Unfortunately, conventional self-supervised learning methods for seq2seq models do not assume that pointer-generator networks are utilized \cite{Song2019Mask,liu2020multilingual,wang2019denoising}.
In practice, the conventional methods are insufficient for pointer-generator networks because they cannot learn to copy tokens from a source text explicitly.

In this paper, we propose a novel self-supervised learning method for pointer-generator networks.
The proposed method, MAsked Pointer-Generator Network (MAPGN), is an extension of MAsked Sequence-to-Sequence pre-training (MASS) \cite{Song2019Mask}.
MASS pre-trains a seq2seq model by predicting the masked tokens taking the masked sequence as input.
In contract, MAPGN can pre-train a copy mechanism efficiently by learning to choose whether to copy or generate tokens with masked tokens.
Our experiments demonstrate that the proposed method is effective for pointer-generator networks with less paired data in two spoken-text normalization tasks, dialect conversion task and spoken-to-written-style conversion task.

\vspace{-2mm}
\section{Pointer-generator networks}
\vspace{-2mm}
This section defines spoken-style normalization with pointer-generator networks.
We define spoken-style text as $\bm{X} =$ $\{x_1, \cdots, x_M\}$ and style normalized text as $\bm{Y} = \{ y_1, \cdots, y_N\}$, 
where $x_m$ and $y_n$ are tokens in the spoken-style and style normalized text, respectively.

\begin{figure}[t]
  \centering
  \centerline{\includegraphics[clip, width=8.5cm]{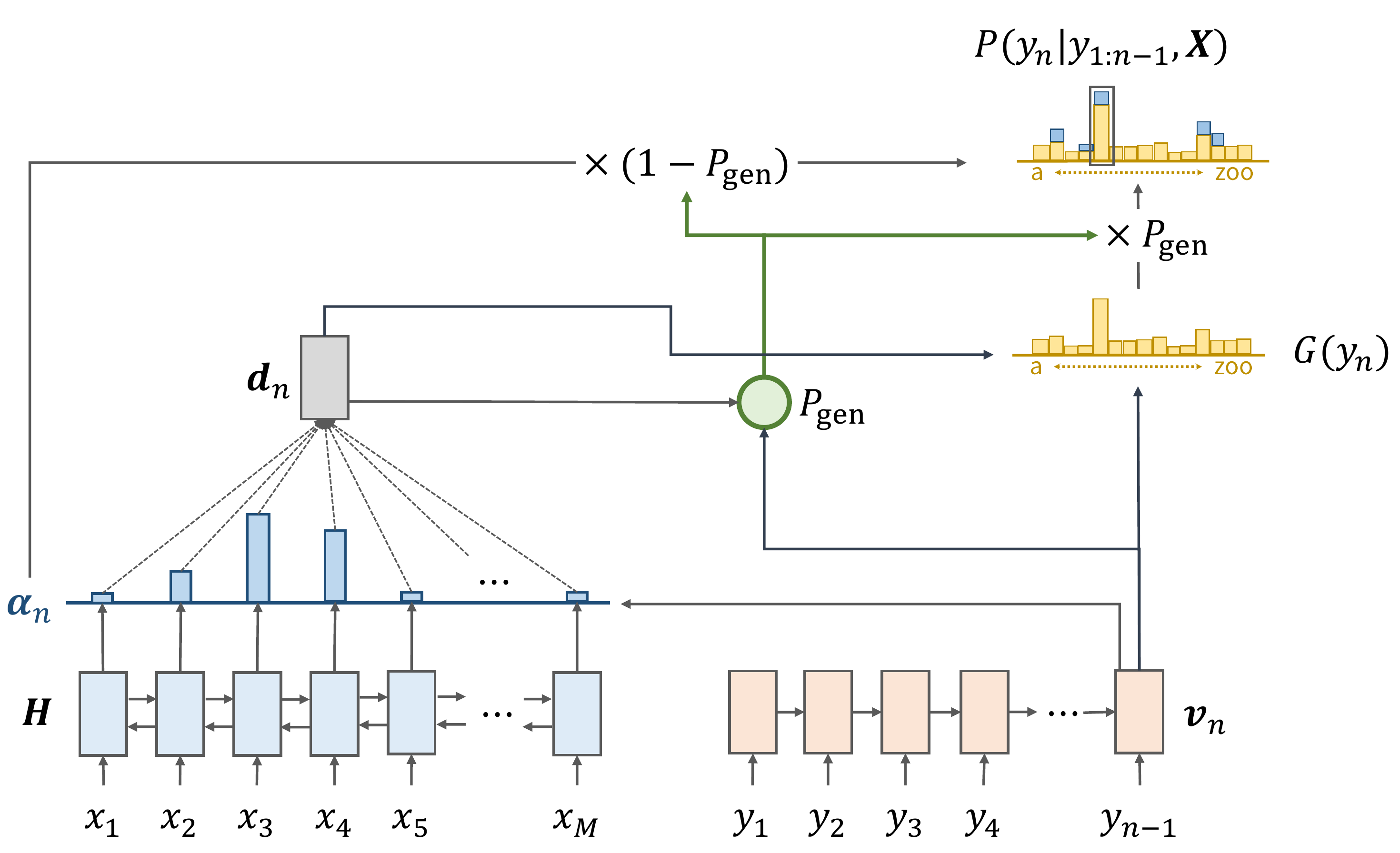}}
  \vspace{-2mm}
  \caption{The network structure of a pointer-generator network.}
  \label{fig:pgn}
  \vspace{-6mm}
\end{figure}

The pointer-generator network predicts generation probabilities of a written-style text $\bm{Y}$ given a spoken-style text $\bm{X}$.
The generation probability of $\bm{Y}$ is defined as 
\begin{equation}
  P(\bm{Y}|\bm{X}; \bm{\Theta}) = \prod_{n=1}^N P(y_n|y_{1:n-1}, \bm{X}; \bm{\Theta}) ,
\end{equation}
where $\bm{\Theta} = \{\theta_{\rm enc}, \theta_{\rm dec}\}$ represents model parameter sets. 
$\theta_{\rm enc}$ and $\theta_{\rm dec}$ are trainable parameter sets with an encoder and a decoder, respectively.
$P(y_n|y_{1:n-1}, \bm{X}; \bm{\Theta})$ can be computed with the encoder and the decoder with a copy mechanism.
Fig. \ref{fig:pgn} shows the network structure of the pointer-generator network.

The encoder converts an input sequence $\bm{X}$ into the hidden representations $\bm{H} = \{\bm{h}_1, \cdots, \bm{h}_M\}$.
These hidden representations are produced by an arbitrary network such as bidirectional recurrent neural networks (RNNs) \cite{schuster1997bidirectional} or a Transformer encoder \cite{vaswani2017attention}.
The decoder computes both copying tokens via pointing and generating tokens from a fixed vocabulary based on a copy mechanism to compute the generation probabilities.
First, the decoder converts tokens from the first token to the $n-1$-th token into hidden vector $\bm{v}_n$.
The hidden vector is produced by an arbitrary network such as unidirectional RNNs \cite{hochreiter1997long} or a Transformer decoder \cite{vaswani2017attention}.
Next, we compute attention distribution $\{\alpha_{1,n}, \cdots, \alpha_{M,n}\}$ from the function that computes the attention distribution with $\bm{H}$ and $\bm{v}_n$.
The attention distribution produces a weighted sum of the encoder states $\bm{d}_n$.
Generation probabilities for the $n$-th token are produced with $\bm{d}_n$ and $\bm{v}_n$ by 
\begin{multline}
  P(y_n|y_{1:n-1}, \bm{X}; \bm{\Theta}) \\
  = P_{\rm{gen}} G(y_n) + (1-P_{\rm{gen}}) \sum_{m:x_m=y_n} \alpha_{m,n},
\end{multline}
\begin{equation}
  G(y_n) = {\tt softmax}({\tt tanh}({[{\bm{d}_n}^{\mathrm{T}}, {\bm{v}_n}^{\mathrm{T}}]}^{\mathrm{T}}; \theta_{\rm dec}); \theta_{\rm dec}),
\end{equation}
\begin{equation}
  P_{\rm{gen}} = {\tt sigmoid}({\tt tanh}({[{\bm{d}_n}^{\mathrm{T}}, {\bm{v}_n}^{\mathrm{T}}]}^{\mathrm{T}}; \theta_{\rm dec}); \theta_{\rm{\rm dec}}),
\end{equation}
where ${\tt tanh}(\cdot)$, ${\tt softmax}(\cdot)$ and ${\tt sigmoid}(\cdot)$ are linear transformational functions with a tanh, softmax and sigmoid activation.
The copy mechanism enables switching probability $P_{\rm{gen}}$ to choose whether to copy or generate tokens.
Thus, $P_{\rm{gen}}$ computes the weighted sum with generator distribution $G(y_n)$ and attention distribution and produces the prediction probability of the $n$-th token.

The model parameter set can be optimized from paired data ${\cal D}_{\rm p}$ $=$ $\{(\bm{X}^1, \bm{Y}^1),$ $\cdots,$ $(\bm{X}^{|{\cal D}_{\rm p}|}, \bm{Y}^{|{\cal D}_{\rm p}|})\}$.
A loss function to optimize the model parameter set is defined as 
\begin{equation}
  {\cal L} = \frac{1}{|{\cal D}_{\rm p}|} \sum_{d=1}^{|{\cal D}_{\rm p}|} \sum_{n=1}^N \log P(y_n^d|y_{1:n-1}^d, \bm{X}^d; \bm{\Theta}). 
\end{equation}

\vspace{-2mm}
\section{Self-supervised learning for pointer-generator networks}
\vspace{-2mm}
This section details self-supervised learning for pointer-generator networks.
The denoising auto-encoder task is widely used for seq2seq self-supervised learning.
In this task, the seq2seq model learns to reconstruct the original text given the corrupted original text by using unpaired data.
The seq2seq model can be fine-tuned with paired data for each subsequent task.
In this paper, we use token span masking which is common with MASS \cite{Song2019Mask} to corrupt the original text for the basic self-supervised learning strategy.
We propose a self-supervised learning method for pointer-generator networks by devising a masking method for the token span masking.

\vspace{-3mm}
\subsection{Basic self-supervised learning strategy}
\begin{figure}[t]
  \centering
  \centerline{\includegraphics[clip, width=8.5cm]{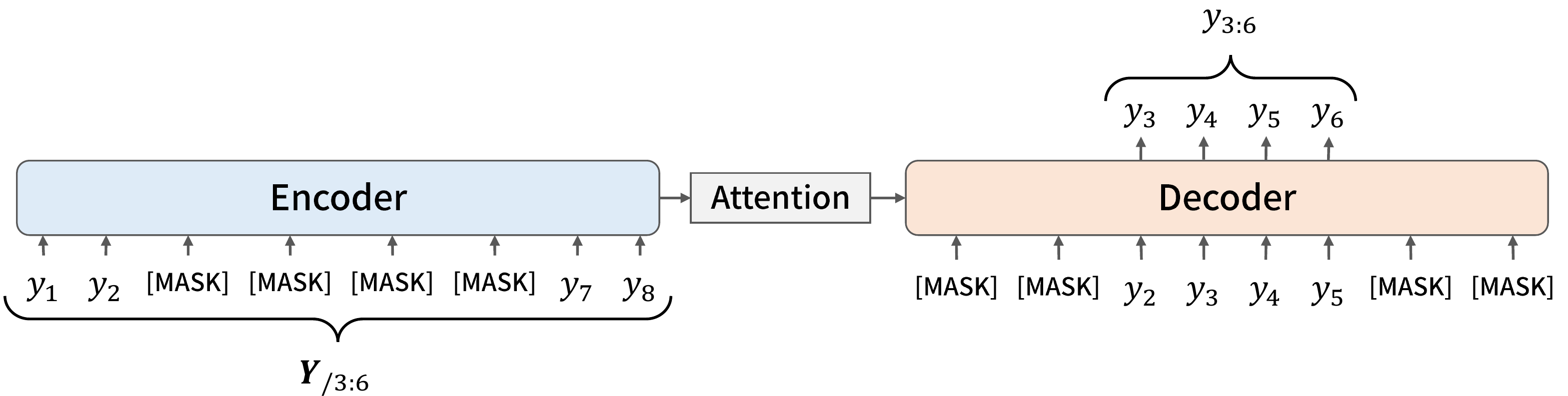}}
  \vspace{-2mm}
  \caption{Token span masking method.}
  \label{fig:mass}
  \vspace{-6mm}
\end{figure}

\vspace{-2mm}
We use token span masking in MASS \cite{Song2019Mask} as a basic self-supervised learning strategy.
In this method, given unpaired sentence $\bm{Y}$, we get $\bm{Y}_{/a:b}$ where its fragment from position $a$ to $b$ are masked.
Here, $0<a<b<n$ and $n$ is the number of tokens of sentence $\bm{Y}$.
The number of tokens that are masked from position $a$ to $b$ is $k=b-a+1$, and the length $k$ is roughly 50\% of $n$.
Position $a$ is selected from between the first token and the $n-k+1$-th token.
$y_{a:b}$ indicates the sentence fragment of $\bm{Y}$ from $a$ to $b$.
In addition, each of the selected masked token is either replaced with a $[{\tt MASK}]$ token, a random token or left unchanged.
We describe details of this replacing method for the masked token in section 3.2.
The seq2seq model is pre-trained by predicting the sentence fragment $y_{a:b}$ taking the masked sequence $\bm{Y}_{/a:b}$ as input, as shown in Fig. \ref{fig:mass}.
The model parameter set can be optimized from unpaired data ${\cal D}_{\rm u}$ $= \{\bm{Y}^1,$ $\cdots,$ $\bm{Y}^{|{\cal D}_{\rm u}|}\}$.
A loss function to pre-train the model parameter set is defined as 
\begin{multline}
  {\cal L} = \cfrac{1}{|{\cal D}_{\rm u}|} \sum_{d=1}^{|{\cal D}_{\rm u}|} \log P(y_{a:b}^d|y_{a-1}^d, \bm{Y}_{/a:b}^d; \bm{\Theta}) \\
           = \cfrac{1}{|{\cal D}_{\rm u}|} \sum_{d=1}^{|{\cal D}_{\rm u}|} \sum_{t=a}^b \log P(y_{t}^d|y_{a-1:t-1}^d, \bm{Y}_{/a:b}^d; \bm{\Theta}).
\end{multline}

In this self-supervised learning, the encoder is encouraged to understand the meaning of unmasked tokens.
Furthermore, the decoder is encouraged to extract more useful information from the encoder side by masking the decoder input tokens which are not masked in the encoder.
If the decoder input tokens are not masked at all, it is assumed that the decoder uses abundant information from the preceding tokens rather than information from the encoder side.
In addition, the decoder can learn more effective language modeling by predicting consecutive tokens in the decoder side rather than predicting discrete tokens.

\vspace{-3mm}
\subsection{Masking methods}
\vspace{-2mm}
We propose a self-supervised learning method for pointer-generator networks.
In token span masking \cite{Song2019Mask}, each of the selected masked token is either replaced with a $[{\tt MASK}]$ token, a random token or left unchanged.
We vary the percentage of these tokens and the replacing method to develop a suitable masking method for pointer-generator networks.
This section describes the conventional method, MASS, two methods that varies the percentage of these replaced tokens in MASS and our proposed method, MAPGN.
% Moreover, we discuss our key idea of MAPGN and the difference between it and MASS in section 3.3.
Table \ref{table:masking_span} summarizes each masking method.

\vspace{-3mm}
\paragraph*{MASS:}
In MASS, 80\% of the masked tokens in the encoder are replaced by $[{\tt MASK}]$ tokens, 10\% are replaced by random tokens, and 10\% are unchanged.
The random tokens are introduced on behalf of the $[{\tt MASK}]$ token, considering that the $[{\tt MASK}]$ token does not appear during fine-tuning.
These tokens are randomly selected from the vocabulary.
In this paper, this masking method is referred to as MASS-1.
Moreover, we prepare two methods that varies the percentage of these replaced tokens: MASS-2 and MASS-3.
In MASS-2, 40\% of the masked tokens in the encoder are replaced by $[{\tt MASK}]$ tokens, 40\% are replaced by random tokens, and 20\% are unchanged.
In MASS-3, 40\% of the masked tokens in the encoder are replaced by $[{\tt MASK}]$ tokens and 60\% are unchanged.

\vspace{-3mm}
\paragraph*{MAPGN:}
Our MAsked Pointer-Generator Network (MAPGN) is an extension of MASS.
In MAPGN, 40\% of the masked tokens in the encoder are replaced by $[{\tt MASK}]$ tokens, 40\% are replaced by random tokens, and 20\% are unchanged.
Key advance of MAPGN is that the random tokens are not selected from all tokens but from tokens in the masking span.
For example, in Fig. \ref{fig:mass}, the random tokens are randomly selected from $\{y_3, y_4, y_5, y_6\}$.
A reason why we use this masking method is detailed in section 3.3.

\vspace{-2mm}
\subsection{Key idea}
\vspace{-2mm}
We presume that MASS-1 is not suitable for pointer-generator network pre-training because the percentage of unchanged tokens in the masking span of the encoder is small and the model cannot learn to copy tokens explicitly.
We assume that pointer-generator networks can learn to copy appropriate tokens from the input by increasing the percentage of unchanged tokens.
Thus, it is assumed that MASS-3 is suitable for pointer-generator network.
However, if only unchanged tokens are increased, there is a possibility that the copy mechanism will be overfitting.
Thus, MAPGN utilizes random tokens to learn to choose to copy or generate tokens effectively.
By comparing MASS-3 with MAPGN, we can verify that increasing unchanged tokens is not enough to pre-train the copy mechanism.
In addition, MAPGN selects random tokens from masking span.
In other words, although random tokens include tokens in the output sequence, these tokens do not have information on appropriate positions.
Thus, these tokens are used to aid for copying tokens and conceal the position of tokens that should be copied.
The role of the random tokens in MAPGN is not only to mask the token but also encourage to determine whether to copy.
By comparing MASS-2 with MAPGN, we can validate the selecting of random tokens from the masking span.

\begin{table}[t]
  \scriptsize
  \centering
  \caption{\label{table:masking_span} Summary of masking methods.}
  \vspace{-2mm}
  \begin{tabular}{|c|cccc|} 
      \hline
       & \multirow{2}{*}{$[{\tt MASK}]$} & \multirow{2}{*}{Random} & \multirow{2}{*}{Unchanged} & Select random \\
       \multicolumn{1}{|c|}{} & \multicolumn{1}{c}{} & \multicolumn{1}{c}{} & \multicolumn{1}{c}{} & \multicolumn{1}{l|}{tokens from} \\
      \hline \hline
      MASS-1 & 80\% & 10\% & 10\% & all tokens \\
      MASS-2 & 40\% & 40\% & 20\% & all tokens \\
      MASS-3 & 40\% & - & 60\% & - \\
      \hline
      MAPGN & 40\% & 40\% & 20\% & masking span \\
      \hline
  \end{tabular}
  \vspace{-4mm}
\end{table}

\vspace{-2mm}
\section{Experiments}
\vspace{-2mm}
This section describes the experimental details of pre-training and fine-tuning on spoken-text normalization tasks.
In particular, we chose dialect conversion and spoken-to-written style conversion tasks in Japanese.
In the dialect conversion task, Japanese dialect is converted into standard Japanese.
In the spoken-to-written style conversion task, spoken-style text produced by an automatic speech recognition system is converted into written-style text with correct punctuations and no disfluencies.

\vspace{-3mm}
\subsection{Datasets}
\vspace{-2mm}
% For the experiments, we prepared following three of datasets.
\paragraph*{Pre-training:}
We prepared a large-scale Japanese web text as the unpaired written-style text data.
The web text was downloaded from various topic web pages using our home-made crawler.
The downloaded pages were filtered in such a way that HTML tags, Javascript codes and other parts that were not useful for these tasks were excluded.
Finally, we prepare {\it one million sentences} for pre-training.

\vspace{-3mm}
\paragraph*{Fine-tuning on dialect conversion task:}
We prepared paired data of a Japanese dialect (Tohoku-ben) and standard Japanese using crowd-sourcing.
We divided the data into a training set, validation set, and test set.
The training set contained 4,492 sentences and we divided the training set into 1,000 and 4,000 sentences to investigate the difference of performance with different amount of training data.
The validation set have 1,702 sentences and the test set have 700 sentences.

\vspace{-3mm}
\paragraph*{Fine-tuning on spoken-to-written style conversion task:}
We used the parallel corpus for Japanese spoken-to-written style conversion (CJSW) \cite{ihori2020parallel}.
Although the CJSW has four domains, we only used the data from one domain (call center dialogue) for training to compare the in-domain (ID) task and the out-of-domain (OOD) task.
The training set has 8,169 sentences, which we divided into increments of 1,000 between 1,000 and 8,000 sentences for the same reason as the dialect conversion task.
A validation set of call center dialogue containing 584 sentences was used for training.
We used all domain test sets: the call center dialogue test set for the ID task and all other test sets for the OOD task.
The test sets were divided in accordance with \cite{ihori2020parallel}.
The datasets were paired data of spoken-style text (manual transcriptions of speech) and written-style text (created by crowd-sourcing).

\vspace{-3mm}
\subsection{Setups}
\vspace{-2mm}
We pre-trained attention-based encoder-decoder networks \cite{luong-etal-2015-effective} and pointer-generator networks \cite{see2017get} with four pre-training methods.
We used the following configurations.
We used pre-trained 512 dimensional word embeddings using continuous bag-of-words \cite{Mikolov2013efficient}.
In the encoder, a 4-layer bidirectional long short-term memory RNN (LSTM-RNN) with 256 units was introduced.
In the decoder, a 2-layer unidirectional LSTM-RNN with 256 units was introduced.
We used an additive attention mechanism \cite{bahdanau2014neural}.
The output unit size (which corresponded to the number of tokens in the training set of word embeddings) was set to 5,640.
To train these networks, we used the adam optimizer and label smoothing with a smoothing parameter of 0.1.
We set the mini-batch size to 64 sentences and the dropout rate in each LSTM-RNN to 0.1.
For the mini-batch training, we truncated each sentence to 200 tokens.
All trainable parameters were randomly initialized.
We used characters as tokens and these pre-trained networks for two tasks in common.

In fine-tuning, we used the attention-based encoder-decoder network and the pointer-generator network which are transferred pre-trained model parameter sets.
We constructed these two networks without pre-training as the baseline.
These model configurations were the same as that of the pre-training model, and all trainable parameters were randomly initialized.
In the evaluation, we calculated automatic evaluation scores in three metrics: BLEU-3 \cite{papineni2002bleu}, ROUGE-L \cite{lin2004automatic}, and METEOR \cite{banerjee-lavie-2005-meteor}.

\begin{table*}[t]
  \scriptsize
  \centering
  \caption{\label{table:dialect} Results of dialect conversion task with 1,000 and 4,000 sentences of training data.}
  \vspace{-2mm}
  \begin{threeparttable}
    \begin{tabular}{|c|rrrr|rrr|} 
        \hline
        \multicolumn{1}{|c}{} & \multicolumn{4}{|c|}{1,000 sentences} & \multicolumn{3}{|c|}{4,000 sentences} \\
        \hline \hline
        Masking method &  & BELU-3 & ROUGE-L & METEOR & BELU-3 & ROUGE-L & METEOR \\
        \hline \hline
        \multirow{2}{*}{baseline} & 1). & 0.112 & 0.350 & 0.241 & 0.441 & 0.618 & 0.523 \\
        & 2). & 0.332 & 0.543 & 0.411 & 0.599 & 0.749 & 0.708 \\
        \hline
        \multirow{2}{*}{MASS-1} & 1). & 0.447 & 0.635 & 0.550 & 0.614 & 0.755 & 0.719 \\
        & 2). & 0.494 & 0.672 & 0.598 & 0.650 & 0.785 & 0.766 \\
        \multirow{2}{*}{MASS-2} & 1). & 0.476 & 0.657 & 0.573 & 0.610 & 0.753 & 0.714 \\
        & 2). & 0.497 & 0.674 & 0.601 & 0.647 & 0.783 & 0.728 \\
        \multirow{2}{*}{MASS-3} & 1). & 0.502 & 0.677 & 0.597 & 0.639 & 0.776 & 0.756 \\
        & 2). & 0.454 & 0.646 & 0.564 & 0.622 & 0.763 & 0.728 \\
        \hline
        \multirow{2}{*}{MAPGN} & 1). & 0.462 & 0.652 & 0.571 & 0.613 & 0.761 & 0.731 \\
        & 2). & \bf{0.535} & \bf{0.704} & \bf{0.646} & \bf{0.656} & \bf{0.789} & \bf{0.770} \\
        \hline
    \end{tabular}
    \begin{tablenotes} \footnotesize
      \item 1). Attention-based encoder-decoder network \hspace{2mm} 2). Pointer-generator network 
    \end{tablenotes} 
  \end{threeparttable}
  \vspace{-2mm}
\end{table*}

\begin{table*}[t]
  \scriptsize
  \centering
  \caption{\label{table:spoken_to_written} Results of spoken-to-written style conversion task with 1,000 and 8,000 sentences of training data.}
  \vspace{-2mm}
  \begin{threeparttable}
    \begin{tabular}{|c|rrr|rr|rr|rr|rr|rr|} 
        \hline
        \multicolumn{1}{|c}{} & \multicolumn{7}{|c|}{1,000 sentences} & \multicolumn{6}{|c|}{8,000 sentences} \\
        \hline \hline
        \multirow{2}{*}{Masking method} & & \multicolumn{2}{c|}{BELU-3} & \multicolumn{2}{|c|}{ROUGE-L} & \multicolumn{2}{|c|}{METEOR} & \multicolumn{2}{|c|}{BELU-3} & \multicolumn{2}{|c|}{ROUGE-L} & \multicolumn{2}{|c|}{METEOR} \\
        & & ID & OOD & ID & OOD & ID & OOD & ID & OOD & ID & OOD & ID & OOD \\
        \hline \hline
        \multirow{2}{*}{baseline} & 1). & 0.177 & 0.061 & 0.360 & 0.247 & 0.296 & 0.235 & 0.651 & 0.245 & 0.736 & 0.412 & 0.809 & 0.394 \\
        & 2). & 0.493 & 0.278 & 0.651 & 0.474 & 0.676 & 0.490 & 0.683 & 0.474 & 0.758 & 0.599 & 0.839 & 0.671 \\
        \hline
        \multirow{2}{*}{MASS-1} & 1). & 0.515 & 0.189 & 0.640 & 0.371 & 0.674 & 0.365 & 0.700 & 0.355 & 0.770 & 0.495 & 0.858 & 0.515 \\
        & 2). & 0.555 & 0.259 & 0.675 & 0.449 & 0.718 & 0.473 & 0.700 & 0.533 & 0.772 & 0.633 & 0.860 & 0.696 \\
        \multirow{2}{*}{MASS-2} & 1). & 0.519 & 0.182 & 0.645 & 0.370 & 0.681 & 0.372 & 0.705 & 0.387 & 0.772 & 0.517 & 0.867 & 0.540 \\
        & 2). & 0.523 & 0.224 & 0.651 & 0.421 & 0.692 & 0.449 & 0.707 & 0.537 & 0.771 & 0.638 & 0.864 & 0.706 \\
        \multirow{2}{*}{MASS-3} & 1). & 0.571 & 0.242 & 0.682 & 0.419 & 0.728 & 0.416 & 0.703 & 0.429 & 0.771 & 0.545 & 0.865 & 0.577 \\
        & 2). & 0.439 & 0.138 & 0.592 & 0.351 & 0.592 & 0.347 & 0.698 & 0.467 & 0.767 & 0.591 & 0.855 & 0.639 \\
        \hline
        \multirow{2}{*}{MAPGN} & 1). & 0.522 & 0.197 & 0.659 & 0.400 & 0.703 & 0.420 & 0.696 & 0.386 & 0.768 & 0.522 & 0.859 & 0.539\\
        & 2). & \bf{0.583} & \bf{0.342} & \bf{0.696} & \bf{0.509} & \bf{0.752} & \bf{0.542} & \bf{0.711} & \bf{0.542} & \bf{0.779} & \bf{0.648} & \bf{0.872} & \bf{0.713} \\
        \hline
    \end{tabular}
    \begin{tablenotes} \footnotesize
      \item 1). Attention-based encoder-decoder network \hspace{2mm} 2). Pointer-generator network 
    \end{tablenotes} 
  \end{threeparttable}
  % \vspace{-1mm}
  \vspace{-6mm}
\end{table*}

\vspace{-3mm}
\subsection{Results}
\begin{figure}[tb]
  \centering
  \centerline{\includegraphics[clip, width=8.0cm]{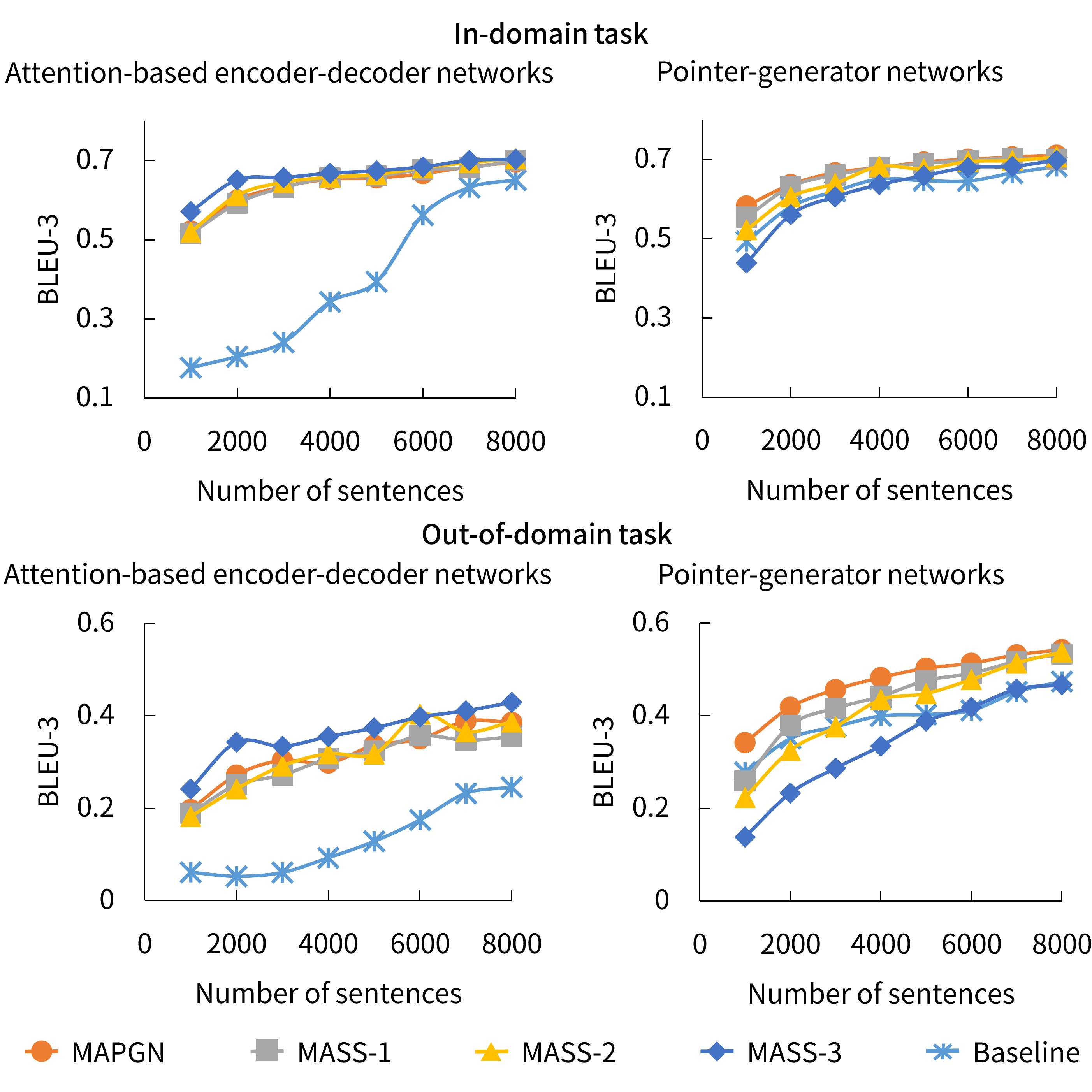}}
  \vspace{-3mm}
  \caption{BLEU-3 score of spoken-to-written style conversion task.}
  \label{fig:spoken_to_written}
  \vspace{-4mm}
\end{figure}

\vspace{-2mm}
Table \ref{table:dialect} and Table \ref{table:spoken_to_written} show the experimental results of the dialect conversion task and the spoken-to-written style conversion task, respectively.
% Table \ref{table:spoken_to_written} shows the results that was fine-tuned with 1,000 and 8,000 sentences of training data.
Fig. \ref{fig:spoken_to_written} shows the BLEU-3 score with all training data in spoken-to-written style conversion task.
Table \ref{table:dialect} shows that MAPGN for pointer-generator networks outperformed other masking methods in all evaluation metrics.
Although MASS-3 was the best performance in encoder-decoder networks, it was the least performance in pointer-generator networks.
The results were the same whether 1,000 or 4,000 sentences were used as training; 
however, the performance of MAPGN improved when 1,000 sentences were used.
Table \ref{table:spoken_to_written} shows that MAPGN for pointer-generator networks outperformed other masking methods in all metrics.
Fig. \ref{fig:spoken_to_written} shows that in encoder-decoder networks, all masking methods improved the baseline performance significantly.
Moreover, MASS-3 was the best performance for any amount of training data.
In pointer-generator networks, MAPGN yield the best performance and the performance was improved more significantly in the OOD task than in the ID task.
In both ID and OOD tasks, the performance difference between each masking method decreased as the amount of paired training data increased.

The results of the two spoken-text normalization tasks largely followed the same trend.
The results of encoder-decoder networks show that MASS-3 outperformed all other methods.
In other words, the pre-training method which learned to actively output the same tokens in the input sequence was more effective.
It is assumed that the method to actively copy tokens is effective for spoken-text normalization tasks using encoder-decoder networks.
Next, a notable point of pre-training in pointer-generator networks results is that each masking method performed differently.
This indicates that the masking method is important for pointer-generator network pre-training.
For example, MASS-3 was the least performance and MASS-2 was less effective than MAPGN.
We can infer that even if the pointer-generator network simply pre-trains to copy tokens actively, the copy mechanism learns to copy tokens that do not need to be copied.
Moreover, we assume that the networks can be pre-trained to copy or not effectively by selecting random tokens from tokens in the masking span rather than from all tokens in the model.
Finally, the results of the pointer-generator networks show that MAPGN outperformed other masking methods.
However, in the ID task, since pointer-generator networks are suitable for spoken-style normalization task and the baseline performance is improved as the amount of paired training data increases, the effectiveness of pre-training for pointer-generator networks decreases.
On the other hand, in the OOD task, pre-training is effective significantly even if the amount of paired training data increases.
Thus, MAPGN is an effective pre-training method for pointer-generator networks if the amount of paired training data is small or in OOD tasks.

\vspace{-2mm}
\section{Conclusion}
\vspace{-2mm}
This paper proposed MAsked Pointer-Generator Network (MAPGN), a self-supervised learning method for pointer-generator networks.
While conventional self-supervised learning methods do not support to explicitly train a copy mechanism in the pointer-generator networks, 
the proposed method can train the copy mechanism efficiently by learning to choose whether to copy or generate tokens against masking span.
Experiments demonstrated that MAPGN outperformed the conventional methods in two spoken-text normalization tasks and was especially effective if the amount of paired training data is small and in OOD tasks.
We concluded that MAPGN is suitable for pre-training pointer-generator networks and effective when paired data set is limited.

\vfill\pagebreak

% \clearpage{
% \footnotesize
% \bibliographystyle{IEEEbib}
% \bibliography{reference}
% }

\end{document}